# "DIVE" into Hydrogen Storage Materials Discovery with AI Agents

Di Zhang*[1], Xue Jia[1], Hung Ba Tran[1], Seong Hoon Jang[1], Linda Zhang[1,2], Ryuhei Sato[3], Yusuke Hashimoto[2], Toyoto Sato[1], Kiyoe Konno[1,4], Shin-ichi Orimo*[1,5], Hao Li*[1]

[1] Advanced Institute for Materials Research (WPI-AIMR), Tohoku University, Sendai 980-8577, Japan

[2] Frontier Research Institute for Interdisciplinary Sciences (FRIS), Tohoku University, Sendai 980-8577, Japan

[3] Department of Materials Engineering, The University of Tokyo, Tokyo 113-8656, Japan

[4] Institute of Fluid Science, Tohoku University, Sendai, 980-8577, Japan

[5] Institute for Materials Research (IMR), Tohoku University, Sendai, 980-8577, Japan

* Emails:

di.zhang.a8@tohoku.ac.jp (D.Z.)

shin-ichi.orimo.a6@tohoku.ac.jp (S.O.)

li.hao.b8@tohoku.ac.jp (H.L.)

**Abstract**

Data-driven artificial intelligence (AI) approaches are fundamentally transforming the discovery of new materials. Despite the unprecedented availability of materials data in the scientific literature, the development of large language model (LLM)-based AI agents that can autonomously transform this knowledge into materials innovation remains limited. Here, we develop the **Descriptive Interpretation of Visual Expression (DIVE) multi-agent workflow**, which systematically reads and organizes experimental data from graphical elements in scientific literatures. Applied to solid-state hydrogen storage materials—a class of materials central to future clean-energy technologies, DIVE markedly improves the accuracy and coverage of data extraction compared to the direct extraction method, with gains of 10–15% over commercial models and over 30% relative to open-source models. Building on a curated database of over 30,000 entries from 4,000 publications, we establish **a rapid inverse-design AI workflow** capable of proposing new materials within minutes. This transferable framework offers a general paradigm for machine-intelligence–driven materials discovery across diverse material classes.

# INTRODUCTION

Data-driven approaches are increasingly reshaping the paradigm of materials discovery and design [1, 2, 3], with the integration of large language models (LLMs) and automated workflows opening new frontiers for accelerated innovation [4, 5, 6]. Central to this vision is the construction of reliable, high-quality materials databases [7]—a persistent challenge that continues to limit the impact of "AI for materials" in both fundamental research and technological deployment. Moreover, rapidly assembling an effective agent or workflow for specific materials problems also remains a substantial barrier [8].

The recent surge in LLM applications has greatly enhanced the prospects for automated data mining and reasoning in materials science. Leveraging advanced LLMs, several studies have explored automated extraction of materials data from scientific literature using prompt engineering and conversational interfaces [9, 10, 11]. Despite these advances, existing strategies still suffer from limitations in completeness, depth, and precision—especially when extracting key quantitative information from graphical elements, which often encode critical materials properties. Current state-of-the-art multimodal models, while powerful, often require multiple rounds of prompt-based querying and validation, resulting in significant computational cost and inefficient use of token resources. There remains a lack of systematic workflows for one-shot, high-throughput extraction and for rigorous, quantitative benchmarking against human-curated data. Moreover, there is no widely adopted workflow for rapidly constructing collaborative, multi-agent materials design systems based on newly mined datasets.

To address these challenges, we present the **Descriptive Interpretation of Visual Expression (DIVE) workflow**—a practical approach that leverages LLMs to recognize figure captions, categorize key graphical content, design targeted prompts, and extract essential descriptive information directly into the text context. This allows the batch extraction of all relevant data points from key figures in a single interaction. Although conceptually simple, **the DIVE pipeline achieves significant gains over current open-source and commercial models, as confirmed by rigorous manual validation and scoring**. Afterward, we apply DIVE to the domain of **solid-state hydrogen storage materials (HSMs)**—a field critical for the future of sustainable, carbon-neutral energy [12]. Hydrogen's high gravimetric energy density and environmentally benign combustion make it an ideal candidate for large-scale energy storage [13], yet practical deployment hinges on the development of compact, safe, and cost-effective storage technologies. Solid-state HSMs, including interstitial hydrides, complex borohydrides, ionic compounds, porous frameworks, and emergent high-entropy and superhydride phases, offer a promising path forward. Despite decades of research, however, no comprehensive, structured experimental database for hydrogen storage materials currently exists.

In this work, we systematically mine over 4,000 primary publications on solid-state HSMs, spanning the period from year of 1972 to 2025, using the DIVE workflow and optimized prompt engineering. Compared to leading multimodal and open-source models, DIVE achieves improvements of 10% to 15% and 30%, respectively, in accuracy and data completeness. The resulting database comprises more than 30,000 entries, which we leverage to construct a materials design agent (***DigHyd***) using GPTs. This agent supports natural language interaction with the HSM database and, more importantly, incorporates a machine-learning-based verifier trained on the extracted data. By integrating LLM-driven reasoning and iterative validation, we realize a streamlined materials design workflow capable of proposing novel hydrogen storage candidates that meet user-defined criteria **within minutes** (**Supplementary Video 1-3**). Overall, this work delivers an efficient, scalable framework for AI-driven materials research and offers a transferable methodology for rapid database construction and inverse design in diverse materials domains.

**RESULTS AND DISCUSSION**

**Figure 1** shows the traditional workflow for extracting materials data from literature using LLMs (**Figure 1a**), as well as the schematic of our proposed DIVE workflow (**Figure 1b**). In the conventional approach, the PDF file of a materials science article is first converted into text (e.g., markdown format) and images. These are then directly fed into a multimodal LLM, which outputs a structured database. In contrast, our DIVE workflow introduces a much more detailed process, particularly for extracting key material properties that are often presented in figures. For HSMs, these include pressure-composition-temperature (PCT) curves, temperature-programmed desorption (TPD) curves, and discharge curves. First, a lightweight inference model scans the article's figure captions to determine whether these key figures are present. If so, the corresponding figure, its caption, and the relevant surrounding text are input into a second multimodal LLM. By carefully designing and optimizing prompts, the LLM is instructed to extract the key points from each curve in the figure, placing the results in the correct positions (as shown in **Figure 1b**, **Prompt**

**Design**). The extracted text then replaces the original figure in the article. We name this approach as the descriptive interpretation of visual expression, as we essentially transform visual information into descriptive text. Finally, the modified article, now with images represented as text, is input into a third LLM for the final extraction of key data (for details on all model combinations, refer to the **Supporting Information**).

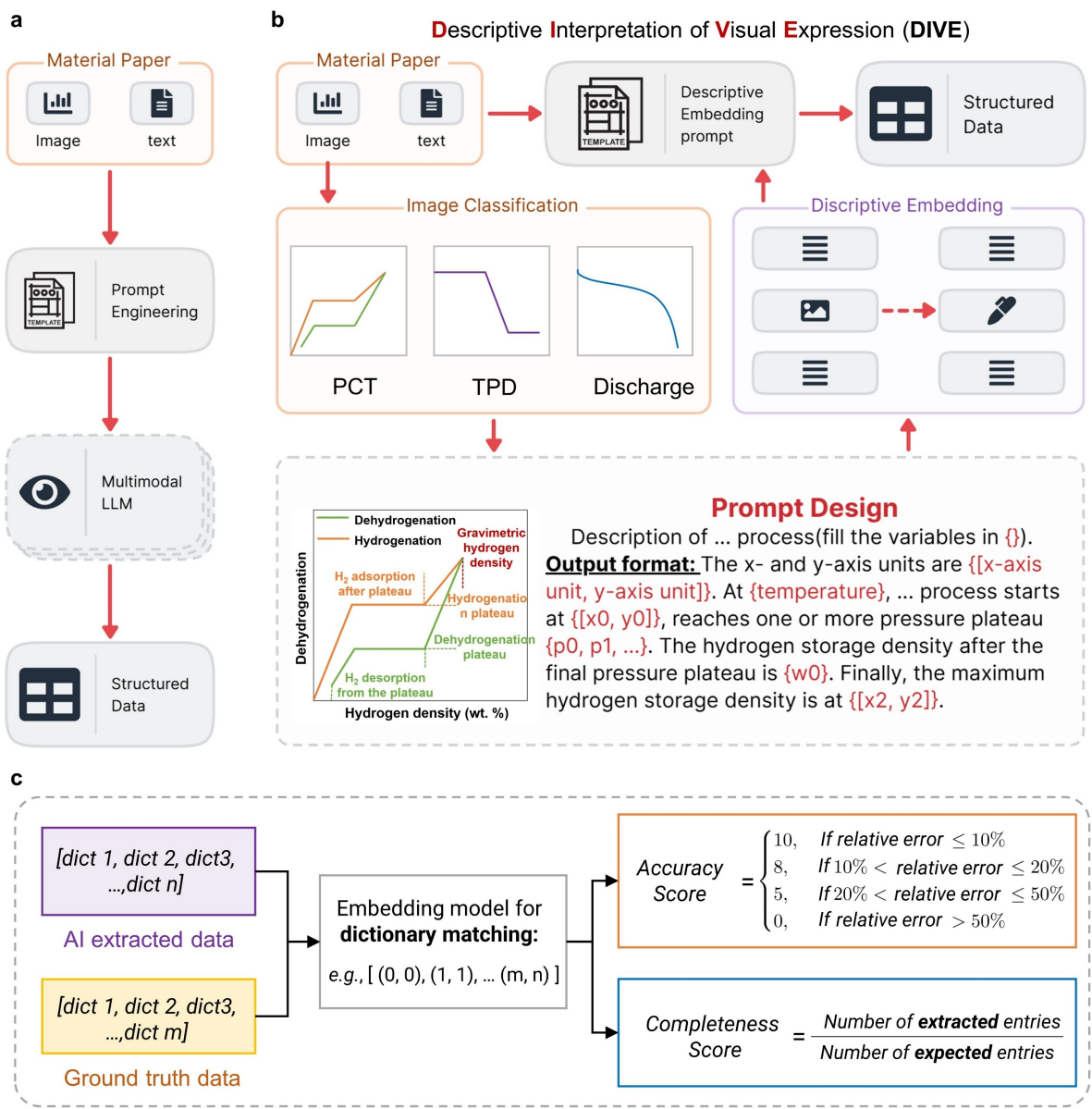

**Figure 1. Schematic diagram and evaluation methods of the DIVE workflow.** (a) Conventional extraction pipeline based on a single multimodal LLM. (b) DIVE extraction pipeline, where descriptive prompts embed key data points and generate image replacements for structured data

extraction. (c) Evaluation method for batch extraction accuracy. Both AI-extracted and manually annotated data are formatted as lists of dictionaries. A shared embedding model is used to match values across dictionaries, from which numerical values are retrieved to calculate precision and completeness scores.

**Equally important to the multi-agent workflow is the development of effective evaluation methods.** To the best of our knowledge, there is currently no well-established method for evaluating the accuracy and completeness of data extraction from articles using LLMs. To save tokens, it is common to extract multiple entries in one call, making the JSON dictionary list format particularly suitable for outputs. However, how to efficiently and reasonably compare human-extracted and AI-extracted JSONs and assign meaningful scores remains underexplored. This is particularly challenging in materials property extraction, where extraction quality cannot be judged simply as true or false, because the magnitude of numerical differences should also be considered. To address this, as shown in **Figure 1c**, we propose using an embedding model to match entries between the human and AI-extracted JSONs. After matching, the units of numerical values are standardized, and the relative errors are calculated using mathematical functions to provide nuanced scoring. We divide the final score into accuracy and completeness (each normalized to 50 points, for a total of 100). This method allows for a more scientific and rapid evaluation of LLM data extraction performance, and can also serve as a reward function for reinforcement learning to further fine-tune or train LLM. The detailed evaluation functions, as well as the code for the DIVE workflow, are available in the **GitHub repository** in the **Data and Code Availability Section** provided with this article. To ensure the high reliability and scientific value of HSMs data, the *DigHyd* Data Checking System (datachecking.dighyd.org, refer to **Supplementary Material** for details) has been developed as an efficient online platform for **manual review and correction of AI-extracted data**.

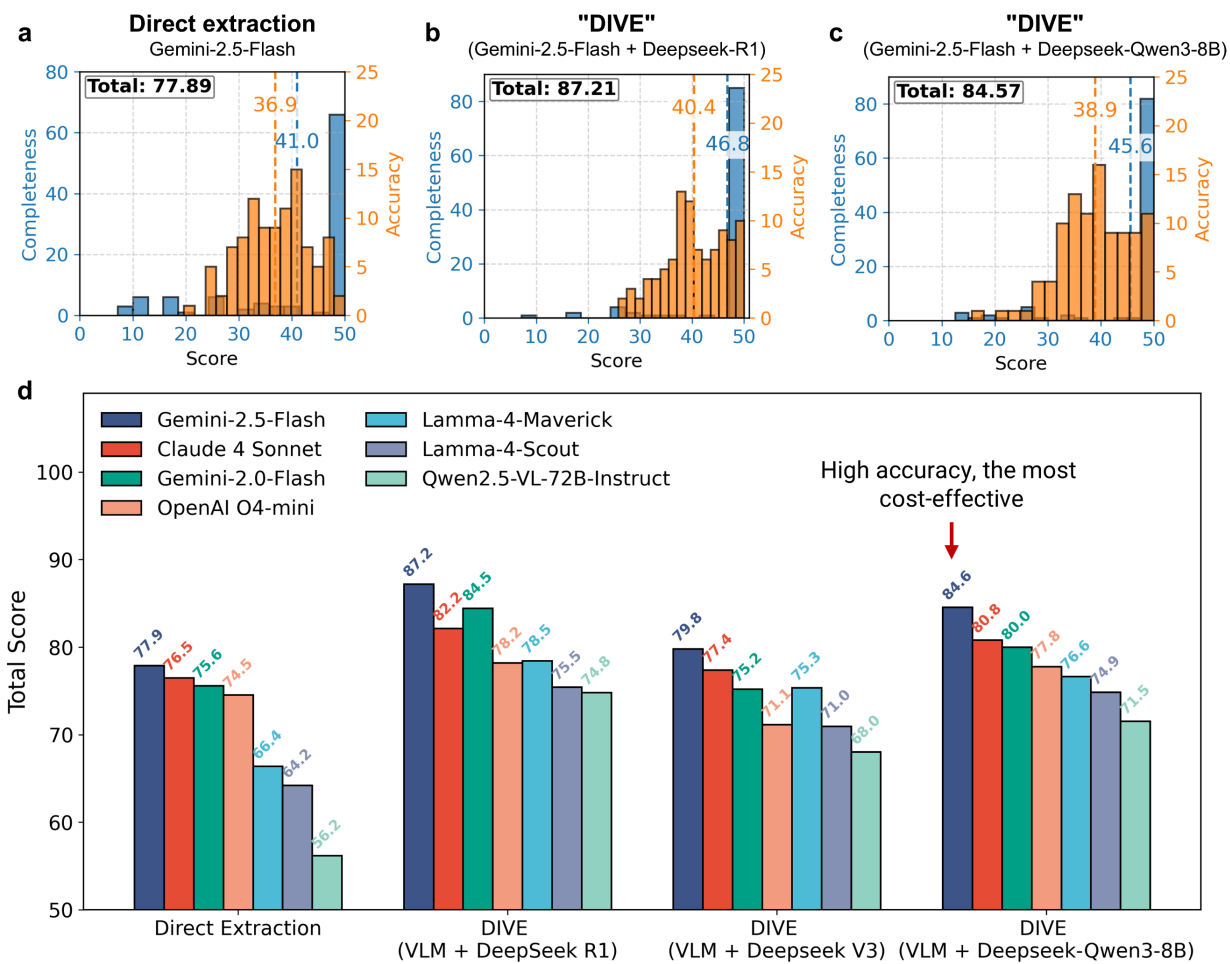

**Figure 2. Performance improvement of the DIVE data extraction workflow.** (a) Conventional extraction workflow using Gemini 2.5 Flash [14]. (b) DIVE workflow integrating Gemini 2.5 Flash with DeepSeek R1. (c) DIVE workflow integrating Gemini 2.5 Flash with DeepSeek Qwen3 8B. (d) Benchmark comparison across seven multimodal models, including four proprietary models (Gemini 2.5 Flash [14], Claude 4 Sonnet [15], OpenAI o4 mini [16], Gemini 2.0 Flash [14]) and three open-source models (LLaMA-4-Scout [17], LLaMA-4-Maverick [17], Qwen2.5-VL-72B-Instruct [18]). Ideally, the proposed DIVE workflow achieves a 10%-15% improvement in extraction performance compared to state-of-the-art commercial models, and an over 30% improvement over leading open-source models.

Based on our developed DIVE workflow and the associated scoring algorithm for materials literature data extraction, we systematically evaluated several state-of-the-art commercial and open-source large language models. The score distributions for data extracted by different combinations of multimodal models and LLMs in the DIVE workflow are benchmarked against a dataset consisting of results manually curated from 100 published articles on experimental HSM reports. **Figure 2a** presents the data extraction scores for the conventional direct extraction approach and under the DIVE workflow (**Figures 2b-c**). Gemini-2.5-Flash [14], currently Google's best model in terms of price-performance, achieved a total score of 77.89 when used for direct extraction. However, when combined in a multi-stage, multi-agent DIVE workflow (Gemini-2.5-Flash[14] + Deepseek R1[19]), the total score increased to 87.21 (**Figure 2b**), representing an improvement of nearly 12%. To further demonstrate the effectiveness of the DIVE workflow on models with even better token efficiency, we also tested DeepSeek-Qwen3-8B [20]. Despite having only 8B parameters, the model also showed about a 10% improvement compared to Gemini-2.5-Flash in the direct extraction scenario. In addition, we systematically assessed the data extraction accuracy across different combinations of mainstream commercial and open-source multimodal and text extraction models (all detailed results can be found in the **Supporting Information**). As shown in **Figure 4d**, for the direct extraction workflow, most commercial models achieved a total score of around 75, whereas open-source models scored noticeably lower. When the multi-stage, multi-agent DIVE workflow is applied—particularly with Deepseek R1 as the post-descriptive embedding LLM—commercial models saw typical improvements of 10%-15%, and open-source models improved by 15%-30%. The highest score was achieved with the combination of Gemini 2.5 Flash and Deepseek R1. However, Deepseek R1 is a large inference model with 685B parameters, making it relatively costly and slow. Therefore, we further tested Deepseek V3 and Deepseek-Qwen3-8B as post-embedding LLMs. Surprisingly, despite its much smaller size (8B parameters), Deepseek-Qwen3-8B achieved a total score of 84.6, second only to the Gemini 2.5 Flash + Deepseek R1 combination, but with much faster inference speed and significantly lower computational cost.

Based on the above benchmarking, we ultimately selected the combination of Gemini 2.5 Flash and Deepseek-Qwen3-8B for data extraction across 4,053 publications. The screening strategy for selecting article DOIs is described in the **Supporting Information**. The processed data have been made publicly available in our **Digital Hydrogen Platform (*DigHyd*: www.dighyd.org)**. **Figure 3** provides an overview of data mining results from over 4,000 hydrogen storage materials publications. As shown in **Figure 3a**, aside from the years before 2010, the number of experimental publications on hydrogen storage materials has steadily increased, with 150–200 papers published annually since 2011 (except for 2021 and 2022, likely due to the global COVID-19 pandemic).

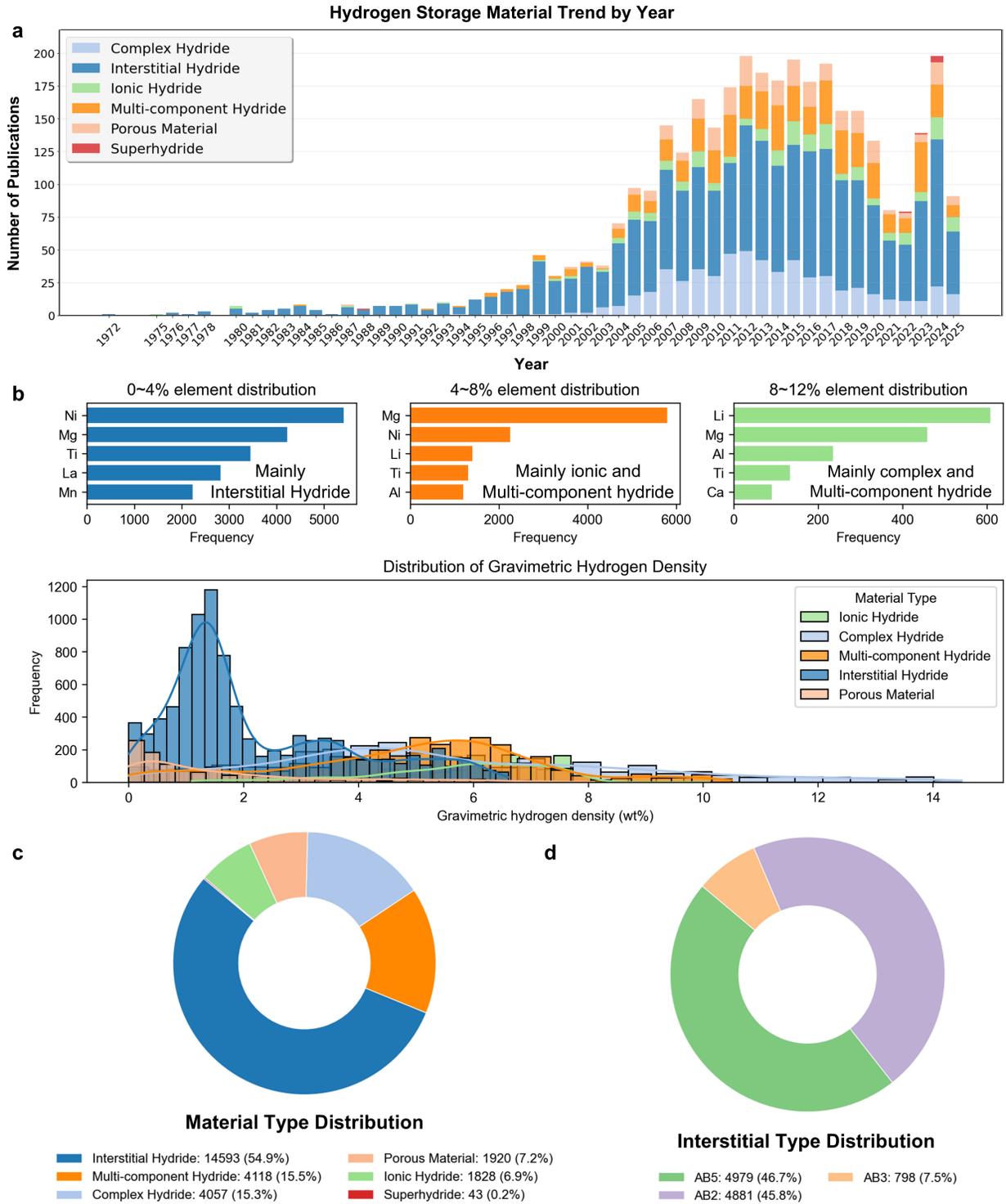

**Figure 3. Overview of data mining from over 4,000 hydrogen storage materials publications.** (a) Annual publication trends categorized by different types of hydrogen storage materials. (b) Distribution of 17,954 hydrogen storage capacity values, along with the elemental distribution of materials within three ranges: 0%-4%, 4%-8%, and 8%-10%. (c) Overall distribution of hydrogen storage material types. (d) Type distribution of interstitial hydrides, classified into $AB_2$, $AB_3$, and $AB_5$ structures.

**Figure 3b** shows the distribution of gravimetric hydrogen densities for different types of hydrogen storage materials. Porous carbon materials generally exhibit very low hydrogen storage capacities at room temperature. Under low temperatures (*e.g.*, 77 K) and moderate pressures (*e.g.*, below 100 bar), their hydrogen uptake is typically in the range of 0–1 wt.%. One of the main advantages of these materials lies in their extremely fast adsorption and desorption kinetics. Therefore, in the hydrogen storage range of 0–1 wt.%, porous materials are the primary candidates [21]. The region with the highest concentration is between 1–2 wt.%, which mainly corresponds to interstitial hydrides—the most widely studied class of hydrogen storage materials. In contrast, ionic, complex, and multi-component hydrides primarily fall in the 4–8 wt.% range. By analyzing the extracted formula fields in the DIVE-generated data dictionaries, we can examine the elemental distribution in hydrogen storage materials across different gravimetric density ranges. The most frequent elements in the 0–4 wt.%, 4–8 wt.%, and 8–12 wt.% intervals are Ni, Mg, and Li, respectively, reflecting a general shift in hydrogen storage materials from interstitial hydrides (represented by $LaNi_5$ [22, 23], Ti-Mn alloys [24], or high-entropy alloys [25]), to ionic hydrides ($MgH_2$), and complex hydrides ($LiBH_4$ [26], $Mg(BH_4)_2$ [27]). **Figures 3c** and **3d** show the proportion of different types of materials in the *DigHyd* platform. Interstitial hydrides account for the largest share, but we also include a small number of superhydrides. Although superhydrides are mainly reported for superconducting applications [28], they are emerging as a new research hotspot for hydrogen storage under ultra-high pressure conditions. **Figure 3d** further illustrates the subtypes of interstitial hydrides.

After constructing the *DigHyd* database, direct data mining enables the extraction of valuable insights for materials design. **Figure 4** illustrates the top five most frequently added elements to typical hydrogen storage materials—$LaNi_5$, $MgH_2$, and $LiBH_4$—and the distribution of key performance metrics for materials modified with these elements. For $LaNi_5$, magnesium is the most commonly used dopant. After Mg is added, the gravimetric hydrogen density of LaNi$_5$-based materials can reach 4-6 wt.% (**Figure 4b**). However, the introduction of Mg also affects the hydrogen absorption and desorption pressures. In the case of MgH$_2$, nickel is the most frequent additive [29]. While doping MgH$_2$ with Ni tends to improve its hydrogen storage density (**Figure 4e**), the dehydrogenation temperature of Mg-Ni systems can reach around 600 K. For LiBH$_4$-based systems, the gravimetric hydrogen density spans the widest range (0-14 wt.%). Notably, introducing carbon or nitrogen can boost the hydrogen density of LiBH$_4$ materials to ~14 wt.%, likely due to the catalytic effects of graphene or N-doped graphene on LiBH$_4$ [30, 31] dehydrogenation. However, despite this high hydrogen density potential, the dehydrogenation temperature of LiBH$_4$ systems also tends to be relatively high, often requiring 700-800 K for complete hydrogen release. All the visualizations shown in **Figures 3-4** can be directly accessed and interacted with *via* our AI agent using natural language (**Supplementary Video 4**).

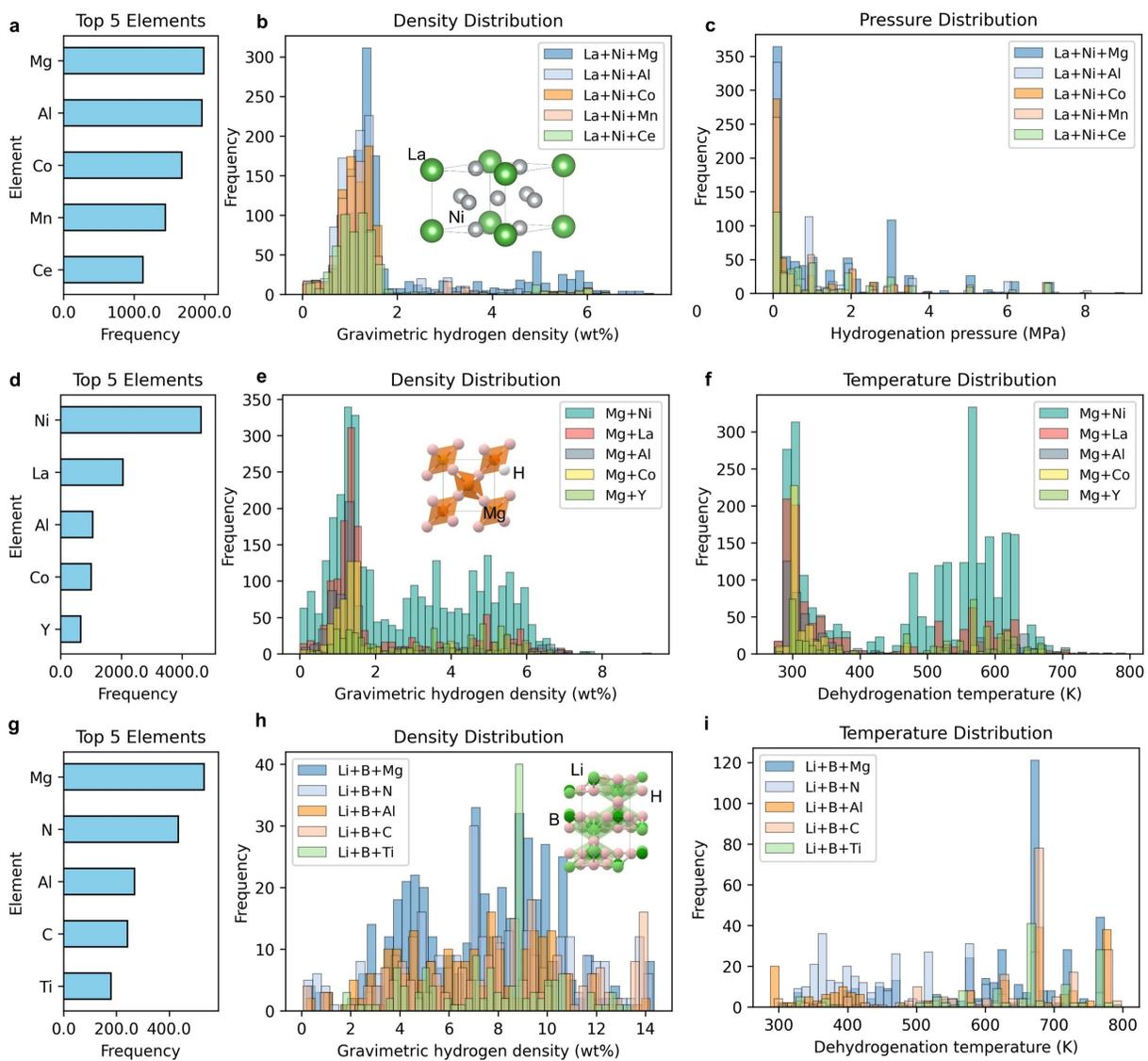

**Figure 4. Analysis of representative hydrogen storage materials.** (a) Top 5 frequently added elements to LaNi$_5$, and the corresponding distributions of (b) hydrogen storage density and (c) hydrogen absorption equilibrium pressure upon element addition; (d) Top 5 frequently added elements to MgH$_2$, and the corresponding distributions of (e) hydrogen storage density and (f) hydrogen desorption temperature; (g) Top 5 frequently added elements to LiBH$_4$, and the corresponding distributions of (h) hydrogen storage density and (i) hydrogen desorption temperature.

Despite decades of research, most HSMs still fall short of the U.S. Department of Energy (DOE) 2030 technical targets for onboard hydrogen storage systems: >5.5 wt% system-level hydrogen capacity, >40 g $H_2$/L volumetric density, operational capability between −40 to 85 °C, and cycling durability exceeding 1,500 charge–discharge cycles [32]. Current benchmark materials exemplify these limitations. $MgH_2$, for instance, boasts a high theoretical gravimetric capacity (7.6 wt%) but requires temperatures above 300 °C for hydrogen release due to slow desorption kinetics [33]. Complex hydrides such as $LiBH_4$ and $NaAlH_4$ can achieve moderate hydrogen densities but often necessitate high temperatures, catalytic activation, or suffer from poor reversibility [34]. Porous frameworks (MOFs/COFs), while tunable and lightweight, rely primarily on weak physisorption and struggle to meet practical storage densities [35]. High-entropy alloys [36] and superhydrides, though scientifically intriguing, demand extreme synthesis or operating conditions (high pressures or cryogenic temperatures) [36, 37], hindering their deployment in commercial systems.

The chemical diversity and complexity of hydrogen storage materials—ranging from $AB_2$, $AB_3$, and $AB_5$ interstitial hydride [38] to Mg-, Ti-, and V-based alloys, complex hydrides, and rare-earth-enriched compounds—make the search for optimal candidates discouraging. Existing efforts to accelerate hydrogen storage material discovery are fragmented. Conventional computational databases primarily focus on crystalline structures and predicted thermodynamic properties, lacking integration with experimentally validated performance data. The absence of a comprehensive, machine-readable platform [39] that integrates both experimental and theoretical information has hindered the rational design and rapid screening of HSMs.

In this work, by integrating the database, machine learning models trained on this database, and LLMs, it becomes straightforward to construct materials-focused AI agents using simple instruction and schema interface functions (for more related details, refer to **Supplementary Information, Figure S9-S11**). To initially assess the reliability of the AI agent's predictions, we did not require *DigHyd* to design entirely new materials. Instead, we focused on cases where comparable materials already exist in the database, allowing for direct validation (**Figure S12** and **Supplementary Video 1**). Under these conditions, the DigHyd agent proposed compositions such as $Mg_2Ni_{0.8}Co_{0.2}$, $Mg_2Fe_{0.8}Co_{0.2}$, and $La_{0.8}Mg_{0.2}Ni_5$. Among these, $Mg_2Fe_{0.8}Co_{0.2}$ was predicted to exhibit a hydrogen storage capacity of 4.06 wt.%. Importantly, analogous alloys already reported in the database, such as $Mg_2FeH_6$ and $Mg_2Fe_{1-x}Co_xH_6$, display capacities in the range of 4.5–5.5 wt.% [40, 41], thereby supporting the consistency of the predictions.

Next, to verify that **DigHyd** can indeed design entirely new materials (**Figure 5** and **Supplementary Video 2**), we applied the same prompting strategy but with explicit instructions to generate compositions never previously reported. Under these conditions, **DigHyd** demonstrated an iterative design–prediction–optimization capability, as illustrated in **Figure 5**. In this workflow, researchers can guide the AI agent to propose novel materials by specifying the material class, potential elements, and target properties such as gravimetric hydrogen density, pressure, and temperature (**Figure 5a**).

In the first round, leveraging the local knowledge base as well as the analytical, reasoning, and predictive capabilities of large language models, the **DigHyd** agent proposed $CaMgFe_2$ (**Figure 5b**). This candidate was then evaluated using our machine learning model, which predicts hydrogen density directly from the material formula. With an $R^2$ value of 0.87, the model provides a reliable first-pass screening for LLM-proposed candidates (**Figure 5c**). $CaMgFe_2$ was predicted to store 2.64 wt.% hydrogen (**Figure 5d**). The agent subsequently suggested increasing the Mg content, resulting in $Mg_2Fe$ with a predicted capacity of 4.13 wt.%. However, literature reports indicated that this compound exhibits hydrogenation/dehydrogenation only at elevated temperatures (300–400 °C), failing to meet the design targets. In response, **DigHyd** refined the composition to $Mg_2Fe_{0.75}Co_{0.25}$, and further to $Mg_2Fe_{0.6}Co_{0.2}Mn_{0.2}$. The latter was predicted to achieve 4.19 wt.% hydrogen storage capacity, with Mn (or alternatively Al) contributing to hydride stabilization and plateau pressure optimization. Importantly, this final composition has never been reported in the current database. Taken together, these results in **Figure 5d** highlight the ability of the **DigHyd** agent to rapidly design, predict, and **iteratively refine candidate materials in line with researcher-defined goals within minutes.** If such AI-driven agents are directly integrated with high-throughput experimental platforms, the efficiency of materials discovery and development could be advanced to an unprecedented level.

To further increase the design difficulty, in the third case study (**Figure S13** and **Supplementary Video 3**), we constrained the element space for material design (A = Mg or Ca, B = Ni). Leveraging the local knowledge base together with the analytical, reasoning, and predictive capabilities of LLM, the **DigHyd** agent proposed 8 candidate materials. Among these, one candidate exceeded the initial target of 4 wt.% hydrogen capacity, while three achieved predicted performances above 3 wt.%. The remaining candidates showed comparatively lower hydrogen densities. Based on these initial predictions, the **DigHyd** agent further optimized the proposed compositions by suggesting minor La and Y doping to enhance hydride phase stability and to reduce the hydrogenation/dehydrogenation temperature and pressure. The final designs, $Mg_2Ni_{2.9}La_{0.1}$ and $Mg_2NiY_{0.1}$, are derived from the $Mg_2Ni$ system, a well-established intermetallic compound for hydrogen storage [42]. The introduction of a small amount of La or Y by partially substituting Ni is a common strategy to optimize hydrogen storage properties. The substitution ratio (3.3% for La [43] or Y [44]) is appropriate because it is sufficient to significantly influence the microstructure and hydrogen storage behavior without destroying the main phase structure. The addition of La or Y can promote grain refinement and introduce defects, which facilitate hydrogen diffusion, improve absorption/desorption kinetics, and may lower the hydrogenation/dehydrogenation temperature. Moreover, the larger atomic radii of La and Y compared to Ni lead to lattice expansion, thus reducing the activation energy for hydrogen diffusion [44]. Therefore, the proposed compositions are also rational for hydrogen storage materials, as supported by both theoretical understanding and experimental data from the literature. In fact, our database did not include this very recent paper [Reference [44]] at the time of writing, which investigates the Mg-Y-Ni system. The findings presented in this work further demonstrate the reliability of the predictions made by our developed agent.

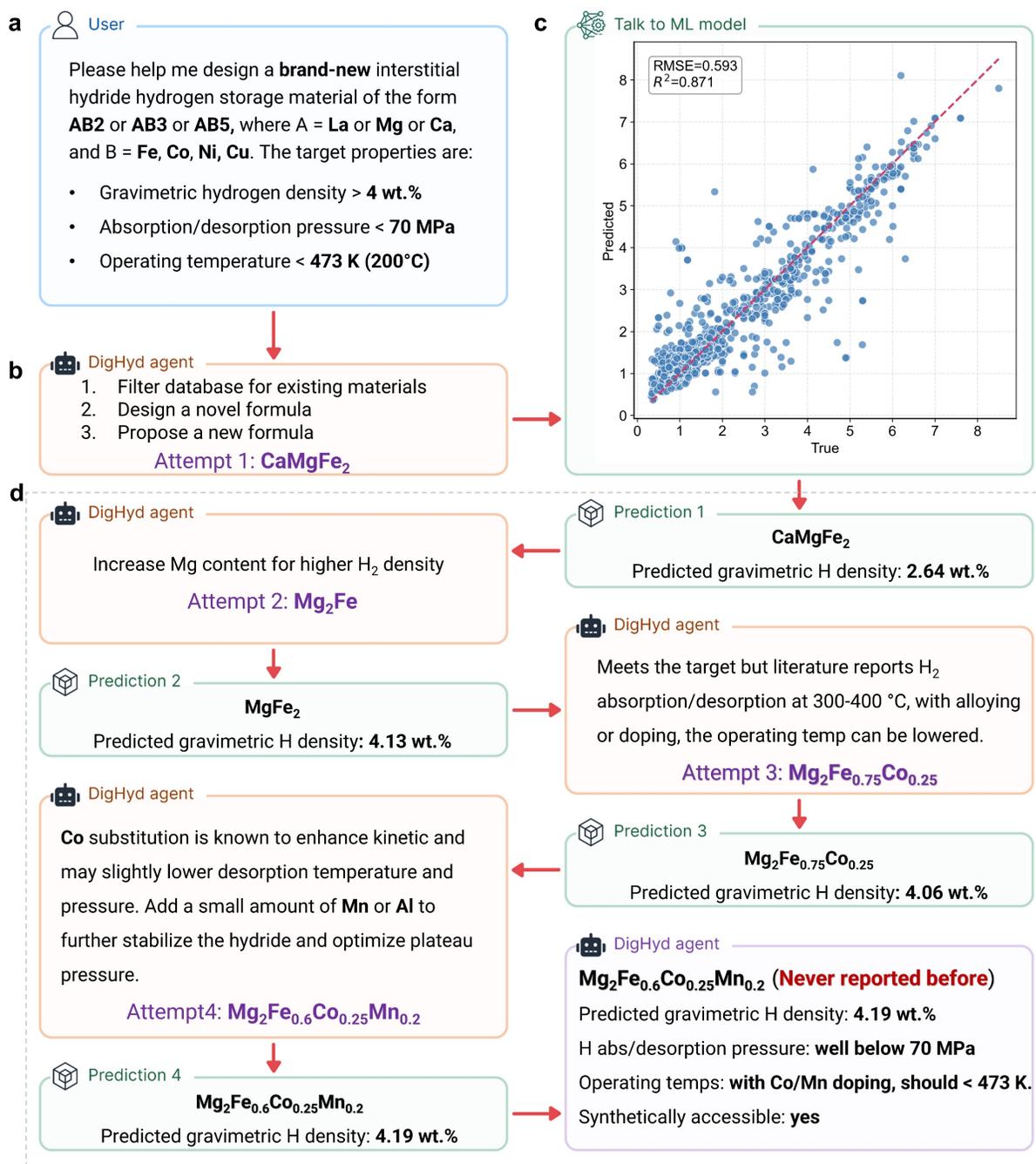

**Figure 5. Workflow of AI agent–driven discovery of new hydrogen storage materials.** (a) The user specifies key requirements, including material type, constituent elements, and performance targets. (b) The *DigHyd* agent proposes initial candidate compositions based on data mined from over 4,000 historical publications. (c) The candidate compositions are evaluated using a pretrained machine learning model to predict their gravimetric hydrogen density. (d) **DigHyd** agent rapidly designs, predicts, and iteratively refines candidate materials in line with researcher-defined goals

within minutes. Finally, the *DigHyd* agent outputs the final material design, together with the relevant reaction conditions and an assessment of synthetic feasibility. (See **Supplementary Video 2** for the complete process and details.)

**METHODS**

**DIVE Workflow**

The first step of the DIVE workflow involves converting PDF files into both text and image formats. This conversion process was accomplished using the MinerU [45], which efficiently extracts both textual content and embedded figures from scientific PDFs. All subsequent steps in the workflow were developed using the LangGraph package, enabling modular and robust pipeline construction for literature mining and data extraction. The complete set of codes including workflow scripts, prompt engineering details, and evaluation protocols, has been made openly available in our GitHub repository (https://github.com/gtex-project/DIVE) to ensure transparency and reproducibility. For the model used in our article (DeepSeek Qwen3 8B), we deployed it locally with an A6000 GPU. For other open- or closed-source models, we accessed them via API calls to third-party platforms or official websites—for example, service providers like SiliconFlow (https://www.siliconflow.com/) and Groq (https://groq.com/).

**The Digital Hydrogen Platform (*DigHyd*) database**

All hydrogen storage materials data extracted via the DIVE workflow have been integrated into the Digital Hydrogen Platform and are accessible through a web interface built with Streamlit (www.dighyd.org). As of August 2, 2025, the database currently contains 4,053 literature sources and 30,435 unique entries, each corresponding to a distinct material or experimental condition. Users can interactively filter data, visualize results, and explore specific material properties or test conditions. We have also deployed the AI agent developed based on DIVE on the website. In addition, the *DigHyd* database is updated daily with newly published literature related to HSMs. The platform also provides direct access to the *DigHyd* agent and integrated machine learning regression models for data analysis and materials prediction.

**Development of the *DigHyd* Agent**

The AI agent utilized in this study was rapidly built using OpenAI's custom GPTs and Actions functionality, allowing seamless integration with local knowledge bases and automated analysis tools. The agent's prompt instructions (**Figure S9-S11**), schema definitions, and action logic are also provided in the GitHub repository (https://github.com/gtex-project/DIVE) for reference and reuse by the community. This infrastructure enables end-to-end question answering, data analysis, and material design based on literature-derived knowledge, supporting both interactive and automated workflows in materials research.

**Machine Learning methods**

We developed a machine learning workflow to predict material properties from chemical composition. After removing samples lacking valid target values or standard chemical formulas, each compound was parsed into the Pymatgen [46] Composition object. A total of 5,357 data points were used in this study. Features were generated using the Matminer ElementProperty featurizer ("magpie" preset) and element molar fractions [47]. The XGBoost regressor was used for prediction, and model performance was evaluated by standard regression metrics. The dataset was randomly split into training and test sets with a ratio of 80%: 20%. Model training was performed using an XGBoost regressor. Hyperparameter optimization was conducted *via* GridSearchCV (with 3-fold cross-validation, scoring by negative mean squared error and parallel computation), to select the best model configuration. Model performance was evaluated using standard regression metrics. All code and scripts are available in our GitHub repository (https://github.com/gtex-project/DIVE).

**Supplementary Materials**

The PDF file includes: Supplementary text, **Figs. S1 to S13**

**Supplementary Video 1:** *DigHyd* agent designs material $Mg_2Fe_{0.8}Co_{0.2}$

**Supplementary Video 2:** *DigHyd* agent designs new material $Mg_2Fe_{0.6}Co_{0.2}Mn_{0.2}$

**Supplementary Video 3:** *DigHyd* agent designs new materials $Mg_2Ni_{2.9}La_{0.1}$ and $Mg_2NiY_{0.1}$

**Supplementary Video 4:** *DigHyd* agent for data analysis and visualization

**Supplementary Video 5:** Main features of the Digital Hydrogen Platform (*DigHyd*)

**Digital Hydrogen Platform (*DigHyd*):** https://www.dighyd.org

**DigHyd Data Checking System:** https://datachecking.dighyd.org

**Code repository:** https://github.com/gtex-project/DIVE

**Acknowledgment**

This work was supported by The Green Technologies of Excellence (GteX) Program Japan Grant No. JPMJGX23H1. We thank Mr. Kazuteru Eguchi and Mrs. Yukina Takada for their efforts in reviewing the data.